\documentclass[journal]{IEEEtran}
\usepackage{graphicx}
\usepackage{cite}

\ifCLASSINFOpdf
\else
\fi
\begin{document}
%
\title{RoTip: A Finger-Shaped Tactile Sensor \\ with Active Rotation}
%
%
%

\author{Xuyang Zhang$^{1,*}$, Jiaqi Jiang$^{1,*}$, and Shan Luo
\thanks{$^{1}$Xuyang Zhang, Jiaqi Jiang and Shan Luo are with the Department of Engineering, King's College London, London WC2R 2LS, U.K. Emails: {\tt\footnotesize \{xuyang.zhang, jiaqi.1.jiang, shan.luo\}@kcl.ac.uk}.}
\thanks{$*$ represents equal contributions.}
}

\maketitle

\begin{abstract}
In recent years, advancements in optical tactile sensor technology have primarily centred on enhancing sensing precision and expanding the range of sensing modalities. To meet the requirements for more skilful manipulation, there should be a movement towards making tactile sensors more dynamic. In this paper, we introduce RoTip, a novel vision-based tactile sensor that is uniquely designed with an independently controlled joint and the capability to sense contact over its entire surface. The rotational capability of the sensor is particularly crucial for manipulating everyday objects, especially thin and flexible ones, as it enables the sensor to mobilize while in contact with the object's surface. The manipulation  experiments demonstrate the ability of our proposed RoTip to manipulate rigid and flexible objects, and the full-finger tactile feedback and active rotation capabilities have the potential to explore more complex and precise manipulation tasks.
\end{abstract}
%
\IEEEpeerreviewmaketitle

\section{Introduction}
%
%
%
%

Tremendous numbers of vision-based tactile sensors have been witnessed in recent years. They can primarily be classified into two main series: the GelSight family~\cite{yuan2017gelsight}, which processes raw images and employs photometric stereo for reconstructing shapes; the TacTip family~\cite{ward2018tactip, zhang2024tacpalm, he2023tacmms}, which tracks the displacement of markers for estimating the contact. Unlike TacTip limited by pin resolution, the GelSight can offer superior resolution and detailed texture information~\cite{yang2017object}. 

Several studies have explored the possibility of implementing rotational movement in tactile sensors. TouchRoller~\cite{cao2023touchroller} was proposed to rapidly reconstruct fabric surfaces via rotating the surface.  TacRot~\cite{zhang2022tacrot}, which utilises a rotating GelSight elastomer, facilitates efficient in-hand object manipulation. Tactile rotation-based gripper~\cite{yuan2023tactile, jiang2024rotipbot} integrates compliant, steerable cylindrical fingertips for improved in-hand manipulation. Nevertheless, they could only perceive the inner side of the fingertip which may limit the interaction mode. To our best knowledge, this is the first tactile sensor that not only can sense the all-around fingertip area, but also actively rotating its body. Those two capabilities are essential to address the challenges of thin and flexible objects, i.e., the incomplete visual perception and the need of dexterity. 


\section{The RoTip sensor}
\subsection{Overview}

\begin{figure*}[ht]
\centering
   \includegraphics[width=1\linewidth]{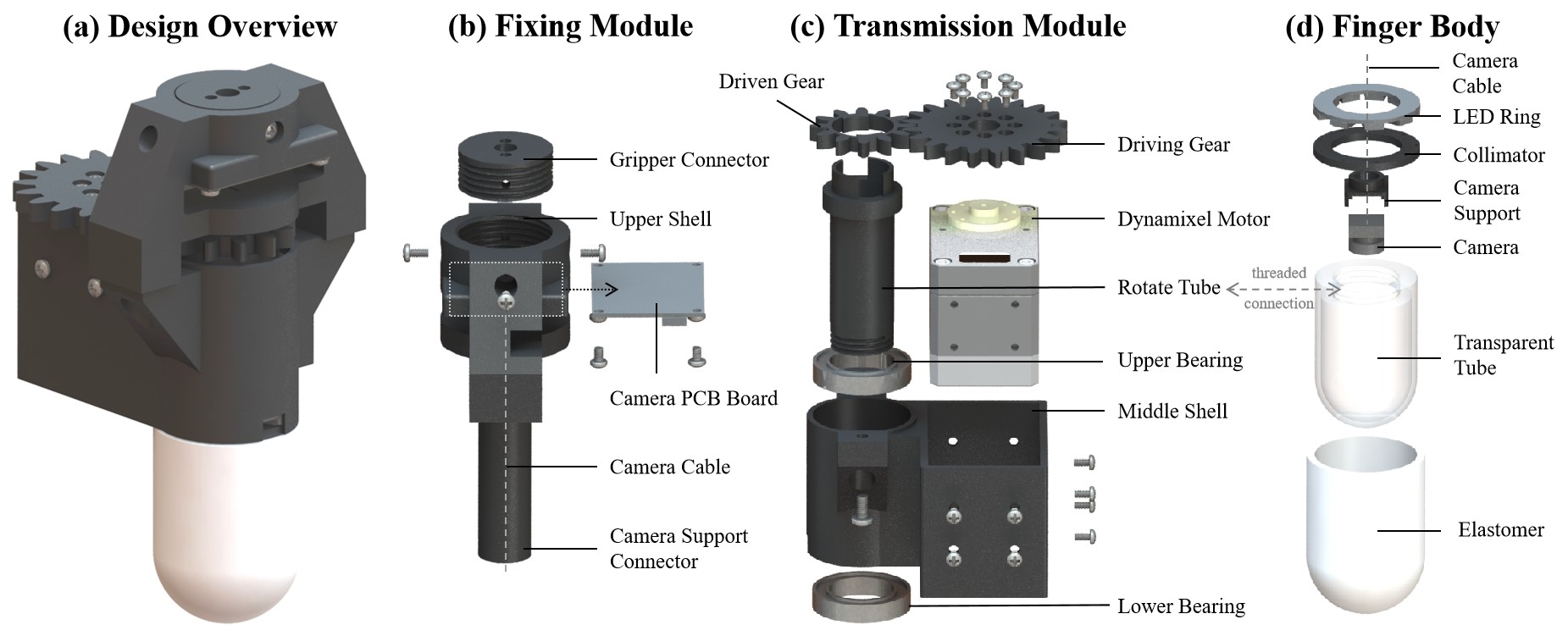}
   \caption{\textbf{From Left to Right:} \textbf{(a) Design overview} of the RoTip sensor, and the exploded view of RoTip's three modules, i.e., \textbf{(b) Fixing module}, \textbf{(c) transmission module} and \textbf{(d) Finger body}. The fixing module is used to connect and provide support to other components of this sensor. The transmission module is used to drive the rotation action enabling dynamic movement. The finger body is the touch interface providing the tactile sensing capability.}
\label{fig: RoTip}
\end{figure*}

There are three main modules of the proposed RoTip sensor: the fixing module, the transmission device and the finger body shown in Fig.~\ref{fig: RoTip}-(b), (c) and (d), respectively. The fixing module is used for secure attachment and precise positioning of the transmission module and the finger body, ensuring stability during operation. Meanwhile, the transmission module is used to drive the rotation action enabling dynamic movement. Lastly, the finger body serves as the touch interface providing the sensor's tactile perception capability.

\subsubsection{Fixing module} 
The fixing module of the RoTip sensor serves to connect and support its components. It features a gripper connector for easy attachment to various robotic grippers, secured with threaded mechanisms and screws to withstand forces and torques during interactions. Additionally, it includes a dedicated support structure for the Printed Circuit Board (PCB) of the camera to ensure stability. The bottom section facilitates camera connection, ensuring its stability during finger body rotation, with the camera cable neatly channeled through a tube within the fixing module to protect it from rotational movements and potential damage.

\subsubsection{Transmission module}
The transmission module of our RoTip sensor employs a gear transmission mechanism. Each finger's Degree of Freedom (DoF) is controlled by a Robotis Dynamixel XM430-W350 actuator, capable of precise position control from 0$^\circ$ to 360$^\circ$ and velocity control for endless turn. A transmission ratio of 5:3 is chosen to balance torque and speed effectively, preventing gear collision when the gripper is closed. Additionally, two 6803-2RS ball bearings (17 mm $\times$ 26 mm $\times$ 5 mm) are incorporated to ensure smooth rotational movement and stability of the system's moving parts.

\subsubsection{Finger body} The finger body is a fingertip-like component used for interaction with the environment or objects. Our RoTip consists of an elastomer with an opaque coating layer crucial for enhancing tactile feedback and a rigid transparent shell that supports the elastomer. Threaded mounting was employed to not only enhance the sensor's stability but also its replicability. This modular approach allows for the easy replacement of the finger body with other shapes, such as one with a flat tip shown in Fig. 2(b) instead of a hemisphere. To enable an even illumination distribution, we designed a printed circuit board (PCB) with multiple LEDs arranged in a specific pattern as shown in Fig. 1(d). To minimise illumination losses, we incorporate a collimator over the LED PCB. Additionally, a diffuser was used to mitigate lighting sparks and to effectively enhance the contrast and uniformity of tactile imprints. 

\begin{figure}[ht]
\setlength{\belowcaptionskip}{-0.5cm}
\centering
  \includegraphics[width=1\linewidth]{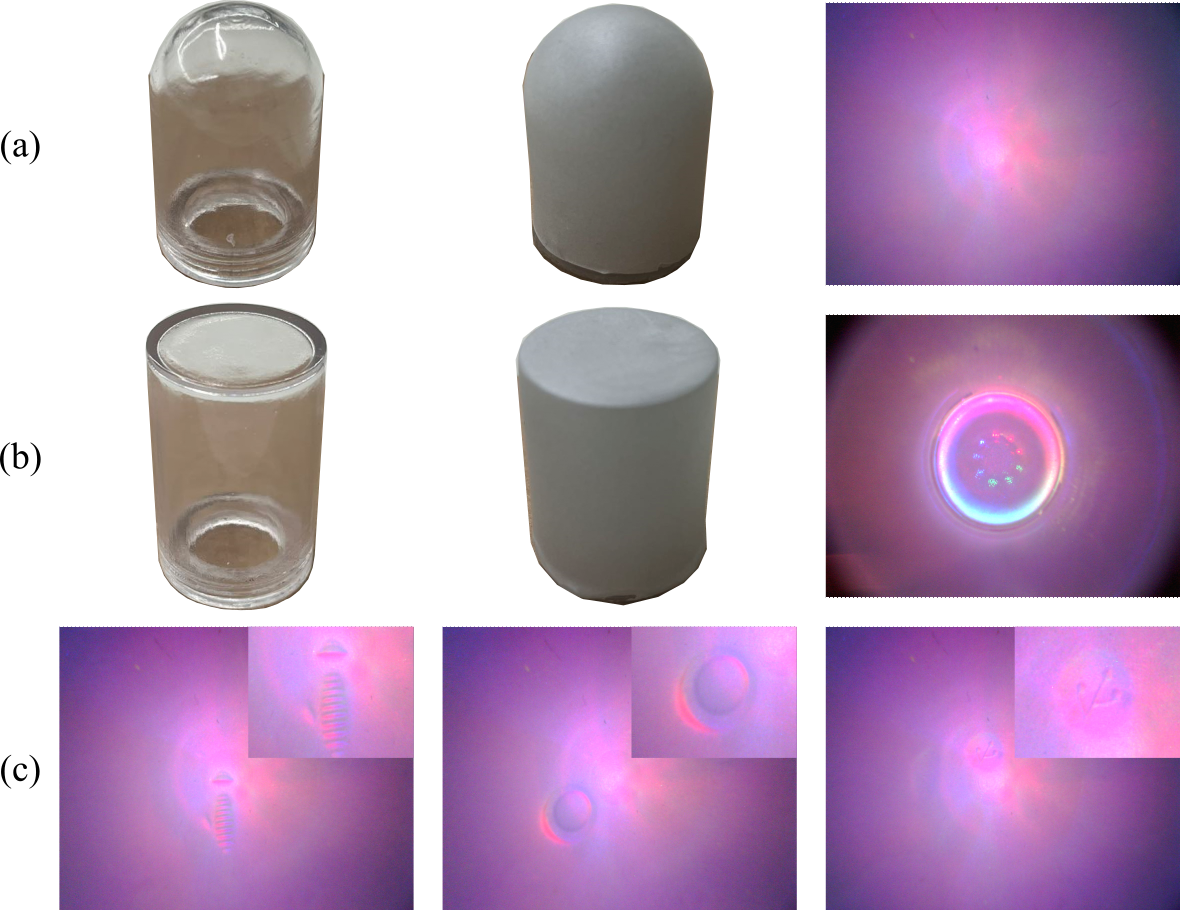}
   \caption{\textbf{(a)} and \textbf{(b)} represent RoTip sensor with a \textbf{hemisphere tip} and a \textbf{flat tip}, respectively. The first two rows display, from left to right, the transparent support shells, the coated elastomer, and the tactile images. Sparking points are noticeable due to the intense reflection from the flat tip. \textbf{(c)} displays the tactile readings captured as the hemisphere-tipped RoTip sensor interacts with a bolt, a nut, and a USB cable.
   }
\label{fig: 3}
\end{figure}

\begin{figure}[ht]
    \centering
    \includegraphics[width = 1\linewidth]{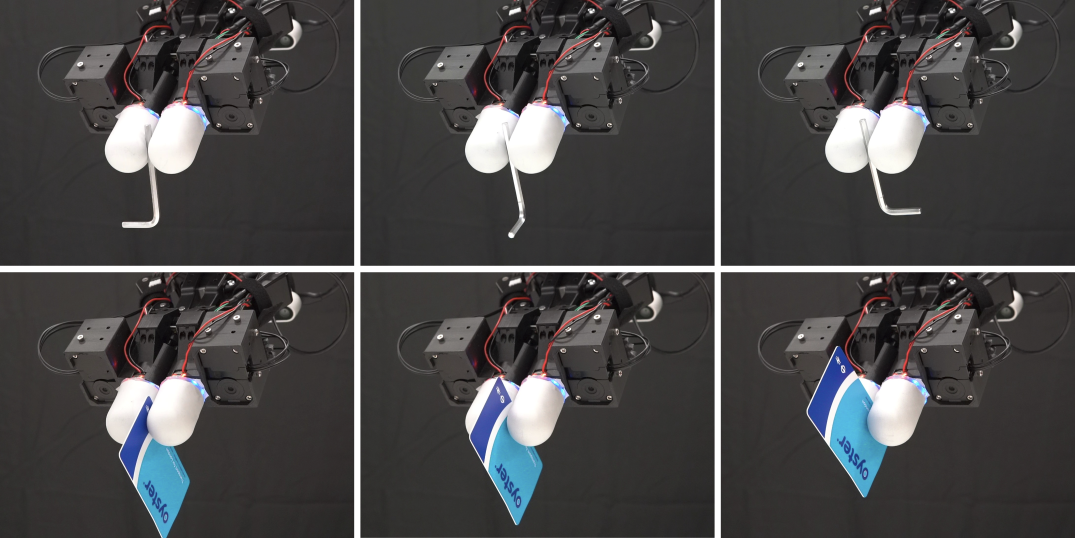}
    \caption{Snapshots of manipulating rigid objects (hex key and oyster card).}
    \label{fig: other applications}
\end{figure}

\section{Manipulation Test}

\begin{figure}[ht]
    \centering
    \includegraphics[width = 1\linewidth]{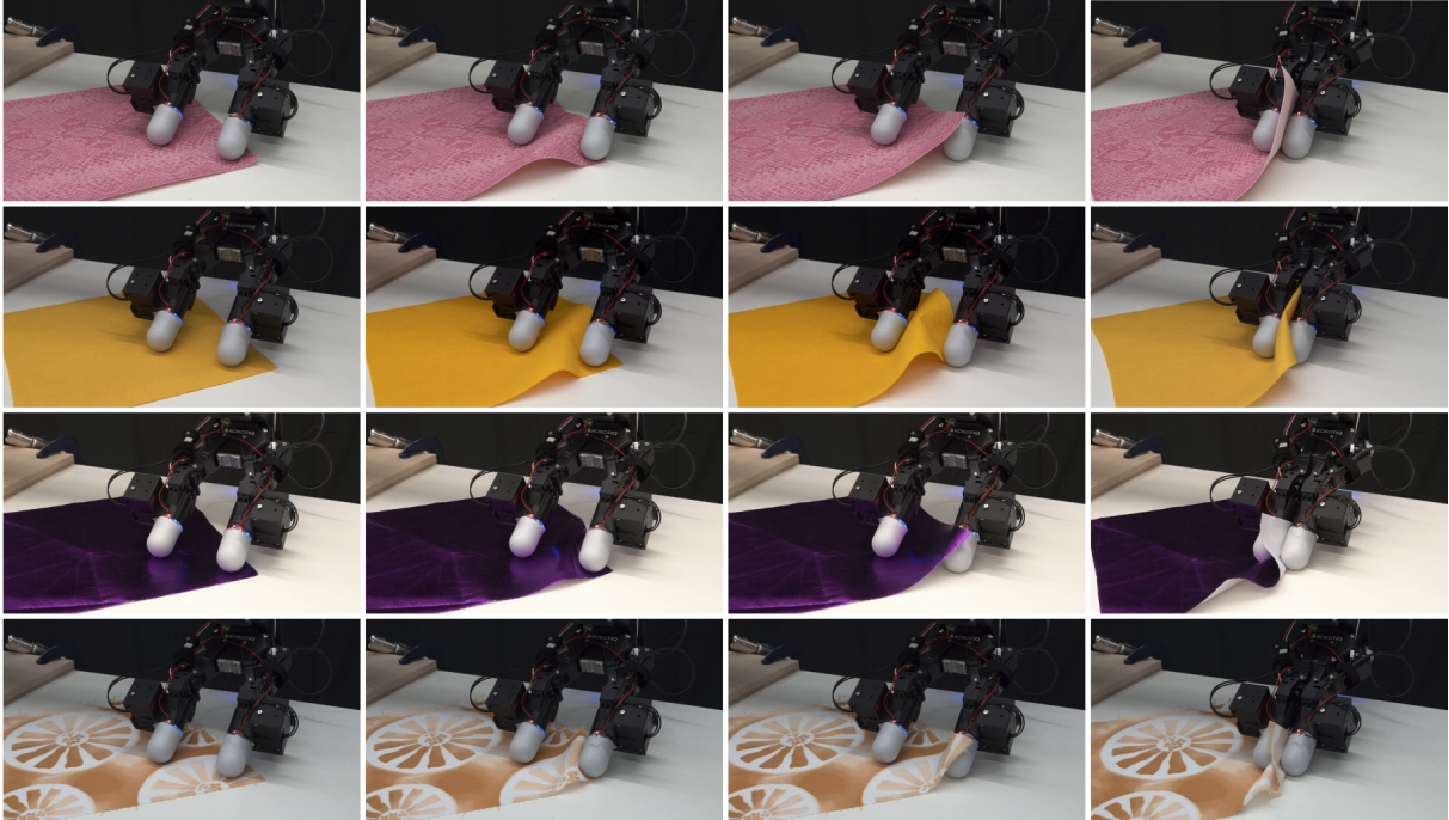}
    \caption{Snapshots of grasping four different fabrics.}
    \label{fig: other applications}
\end{figure}

We first validated the manipulation capability of the parallel gripper integrated with the proposed fingers on rigid objects, as shown in Fig. 3. When both RoTip fingers rotate clockwise, they induce the hex key to spin along its vertical axis. Conversely, when they rotate in opposite directions, they manipulate the oyster card through translational movement. 

We then evaluated the grasping capability of the gripper on the thin and flexible objects, shown in Fig. 4. The gripper first moves to the corner position of the fabric, with one finger pressing down on the material, while the other finger rotates to squeeze the fabric's corner. Subsequently, as the fabric's corner bends under the rotational movement of the fingers, it is rolled between the two fingers. Then, the gripper fingers close, completing the grasp of the fabric corner.


%

\ifCLASSOPTIONcaptionsoff
  \newpage
\fi

\bibliographystyle{unsrt}
\bibliography{main}

\end{document}